\def\year{2019}\relax
\documentclass[letterpaper]{article} 
\usepackage{aaai19}  
\usepackage{times}  
\usepackage{helvet}  
\usepackage{courier}  
\usepackage{url}  
\usepackage{graphicx}  

\usepackage{booktabs}       
\usepackage{amsfonts}       
\usepackage{nicefrac}       
\usepackage{microtype}      
\usepackage{subfigure}
\usepackage{caption}
\usepackage{subfigure}
\usepackage{amssymb}
\usepackage{todonotes}
\usepackage{caption}

\usepackage{lipsum}
\newcommand{\ie}{\emph{i.e., }}
\newcommand{\eg}{\emph{e.g., }}

\begin{document}
%
\title{Explicit Interaction Model towards Text Classification}
\author{
Cunxiao Du\textsuperscript{1},
Zhaozheng Chen\textsuperscript{1}\thanks{Equal contribution.},
Fuli Feng\textsuperscript{2}\footnotemark[1],
Lei Zhu\textsuperscript{3},
Tian Gan\textsuperscript{1}\thanks{Corresponding Author.}, 
Liqiang Nie\textsuperscript{1}\\
\textsuperscript{1}Shandong University, No.72 Binhai Road, Jimo,Qingdao, Shandong, China 266237\\
\textsuperscript{2}National University of Singapore, 13 Computing Drive, Singapore 117417\\
\textsuperscript{3}Shandong Normal University,
No.1 University Road, Changqing Dist.,
Ji'nan, Shandong, China 250358
\\
\{cnsdunm,zhaozhengcc,fulifeng93,leizhu0608\}@gmail.com,
\{gantian, nieliqiang\}@sdu.edu.cn
}


\maketitle

\begin{abstract}
Text classification is one of the fundamental tasks in natural language processing. Recently, deep neural networks have achieved promising performance in the text classification task compared to shallow models. Despite of the significance of deep models, they ignore the fine-grained (matching signals between words and classes) classification clues since their classifications mainly rely on the text-level representations. To address this problem, we introduce the interaction mechanism to incorporate word-level matching signals into the text classification task. In particular, we design a novel framework, EXplicit interAction Model (dubbed as EXAM), equipped with the interaction mechanism. We justified the proposed approach on several benchmark datasets including  both multi-label and multi-class text classification tasks. Extensive experimental results demonstrate the superiority of the proposed method. As a byproduct, we have released the codes and parameter settings to facilitate other researches. 
\end{abstract}

\section{Introduction}
Text classification is one of the fundamental tasks in natural language processing, targeting at classifying a  piece of text content into one or multiple categories. According to the number of desired categories, text classification can be divided into two groups, namely, \textit{multi-label} (multiple categories) and \textit{multi-class} (unique category). For instance, classifying an article into different topics (\eg machine learning or data mining) falls into the former one since an article could be under several topics simultaneously. By contrast, classifying a comment of a movie into its corresponding rating level lies into the multi-class group. Both multi-label and multi-class text classifications have been widely applied in many fields like sentimental analysis \cite{SenticNet}, topic tagging \cite{fasttext}, and document classification \cite{HAN}.

\textit{Feature engineering} dominates the performance of traditional shallow text classification methods for a very long time. Various rule-based and statistical features like bag-of-words \cite{bow} and N-grams \cite{ngram} are designed to describe the text, and fed into the shallow machine learning models such as Linear Regression \cite{LR} and Support Vector Machine \cite{Cortes1995SupportvectorN} to make the judgment. Traditional solutions suffer from two defects: 1) High labor intensity for the manually crafted features, and 2) data sparsity (a N-grams could occur only several times in a given dataset). 

Recently, owing to the ability of tackling the aforementioned problems, deep neural networks \cite{TextCNN,dan,vdcnn,TextRNN,fasttext} have become the promising solutions for the text classification. Deep neural networks typically learn a \textit{word-level representation} for the input text, which is usually a matrix with each row/column as an embedding of a word in the text. They then compress the word-level representation into a \textit{text-level representation} (vector) with aggregation operations (\eg pooling). Thereafter, a fully-connected (FC) layer at the topmost of the network is appended to make the final decision. Note that these solutions are also called \textit{encoding-based methods} \cite{encoder}, since they encode the textual content into a latent vector representation.

\begin{figure}[htbp] \centering    
	\includegraphics[width=0.9\columnwidth]{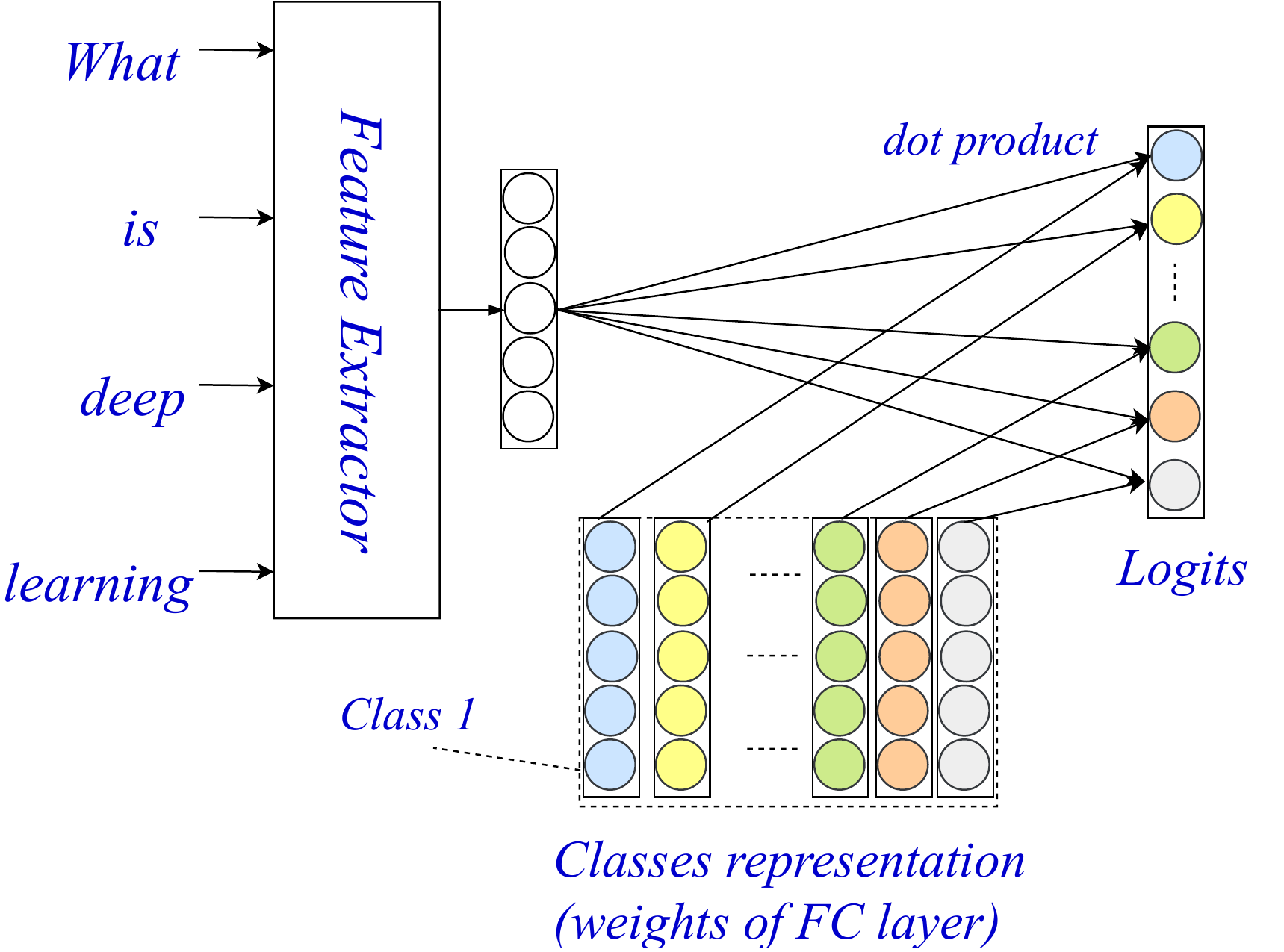} 
	\caption{Illustration of encoding-based methods for text classification with text-level matching.}     
	\label{dense_intro}
\end{figure}

Although great success has been achieved, these deep neural network based solutions naturally ignore the fine-grained classification clues (\ie matching signals between words and classes), since their classifications are based on text-level representations. As shown in Figure \ref{dense_intro}, the classification (\ie FC) layer of these solutions matches the text-level representation with class representations via a dot-product operation. Mathematically, it interprets the parameter matrix of the FC layer as a set of class representations (each column is associated with a class) \cite{output}. As such, the probability of the text belonging to a class is largely determined by their overall matching score regardless of word-level matching signals, which would provide explicit signals for classification (\eg \textit{missile} strongly indicates the topic of \textit{military}).

To address the aforementioned problems, we introduce the interaction mechanism \cite{llstm}, which is capable of incorporating the word-level matching signals for text classification. The key idea behind the interaction mechanism is to explicitly calculate the matching scores between the words and classes. From the word-level representation, it computes an interaction matrix, in which each entry is the matching score between a word and a class (dot-product between their representations), illustrating the word-level matching signals. By taking the interaction matrix as a text representation, the later classification layer could  incorporate fine-grained word level signals for the finer classification rather than simply making the text-level matching.

Based upon the interaction mechanism, we devise an EXplicit interAction Model (dubbed as \textit{EXAM}). Specifically, the proposed framework consists of three main components: \textit{word-level encoder}, \textit{interaction layer}, and \textit{aggregation layer}. The word-level encoder projects the textual contents into the word-level representations. Hereafter, the interaction layer calculates the matching scores between the words and classes (\ie constructs the interaction matrix). Then, the last layer aggregates those matching scores into predictions over each class, respectively. We justify our proposed EXAM model over both the multi-label and multi-class text classifications. Extensive experiments on several benchmarks demonstrate the effectiveness of the proposed method, surpassing the corresponding state-of-the-art methods remarkably. 

In summary, the contributions of this work are threefold: 
\begin{itemize}
\item  We present a novel framework, EXAM, which leverages the interaction mechanism to explicitly compute the word-level interaction signals for the text classification.
\item We justify the proposed EXAM model over both multi-label and multi-class text classifications. Extensive experimental results demonstrate the effectiveness of the proposed method.
\item We release the implementation of our method (including some baselines) and the involved parameter settings to facilitate later researchers\footnote{https://github.com/NonvolatileMemory/AAAI\_2019\_EXAM .}.
\end{itemize}

\section{Preliminaries}
In this section, we introduce two widely-used word-level encoders: \textit{Gated Recurrent Units} \cite{GRU} and \textit{Region Embedding} \cite{regionemb}. These encoders project a piece of input text into a word-level representation, serving as the building blocks of the proposed method. 
For the notations in this paper,
we use bold capital letters (e.g., \textbf{X}) and bold lowercase letters (e.g., \textbf{x}) to denote matrices and vectors, respectively. We employ non-bold letters (e.g., x) to represent scalars, and Greek letters (e.g., $\alpha$ ) as parameters. $\mathbf{X_{i,:}}$ is used to refer the i-th row of the matrix $\mathbf{X}$, $\mathbf{X_{:,j}}$ to represent the j-th column vector and $\mathbf{X_{i,j}}$ to denote the element in the i-th row and j-th column. 

\subsection{Gated Recurrent Units} 
Owing to the ability of capturing the sequential dependencies and being easily optimized (\ie avoid the gradient vanishing and explosion problems), Gated Recurrent Units (GRU) becomes a widely used word-level encoder \cite{TextRNN,dlstm}. 
Typically, a GRU generates word-level representations in two phases: 1) mapping each word in the text into an embedding (a real-valued vector), and 2) projecting the sequence of word embeddings into a sequence of hidden representations, which encodes the sequential dependencies.

\noindent \textbf{Word embedding.}
Word embedding is a general method to map a word from one hot vector to a low dimensional and real-valued vector. With enough data, word embedding can capture high-level representations of words.
\\
\textbf{Hidden representation.}
Given an embedding feature sequence $\mathbf{E = [E_{1,:}, E_{2,:}, \cdot \cdot \cdot, E_{n,:}] }$, GRU will compute a vector $\mathbf{H_{i,:}}$ at the i-th time-step for each $\mathbf{E_{i,:}}$, and  $\mathbf{H_{i,:}}$ is defined as:
\begin{equation}
\left\{
\begin{array}{lr}
\mathbf{r_{i} = \sigma(M_{r}\cdot[H_{i-1,:},E_{i,:}])},\\
\mathbf{z_{i} = \sigma(M_{z}\cdot[H_{i-1,:},E_{i,:}])},\\
\mathbf{\widetilde{H_{i,:}} = tanh(M_{r}\cdot[H_{i-1,:},E_{i,:}])},\\
\mathbf{H_{i,:} = (1-z_{i})*H_{i-1,:}+z_{i}*\widetilde{H_{i,:}}},
\end{array}
\right.
\end{equation}
where $\mathbf{M_{r}}$ and $\mathbf{M_{z}}$ are trainable parameters in the GRU, and $\sigma$ and $tanh$ are sigmoid and tanh activation functions, respectively.
The sequence of hidden representations $\mathbf{H = [H_{1,:}, \cdots, H_{n,:}]}$ is denoted as the word-level representation of the input text.

\subsection{Region Embedding}
Although word embedding is a good representation for the word, it can only compute the feature vector for the single word. Qiao et al. \shortcite{regionemb} proposed region embedding to learn and utilize task-specific distributed representations of N-grams. In the region embedding layer, the representation of a word has two
parts, the embedding of the word itself and a weighting matrix to interact with the local context. For the word $w_{i}$, the first part $\mathbf{e_{w_{i}}}$ is learned by an embedding matrix $\mathbf{E} \in \mathbb{R}^{k \times v}$ and the second part $\mathbf{K_{w_{i}}} \in \mathbb{R}^{k\times(2\times s+1)}$ is looked up in the tensor $\mathbf{U} \in \mathbb{R}^{k\times(2 \times s + 1)\times v}$ by $w_{i}$'s index in the vocabulary, where $v$ is the size of the vocabulary, $2 \times s + 1$ the region size and $k$ the embedding size. And then, each column in $\mathbf{K_{w_{i}}}$ is used to interact with the context word in the corresponding relative position of $w_{i}$ to get the context-aware $\mathbf{p^{t}_{w_{i+t}}}$ for each word $w_{i+t}$ in the region. Formally it is computed by the following function:
\begin{equation}
\mathbf{p^{i}_{w_{i+t}} = K_{w_{i},t} \odot \mathbf{e_{w_{i+t}}}},
\end{equation}
where $\odot$ denotes element-wise multiply.
And the final representation $\mathbf{r_{i,s}}$ of the middle word $w_{i}$ is computed as follows:
\begin{equation}
\mathbf{r_{i,s}} = max\mathbf{([p^{i}_{w_{i-s}}, p^{i}_{w_{i-s+1}}, \cdot \cdot \cdot, p^{i}_{w_{i+s-1}}, p^{i}_{w_{i+s}}])}.
\end{equation}

\section{Model}
\subsection{Problem Formulation}
\begin{itemize}
\item Multi-Class Classification. 
In this task, we should categorize each text instance to precisely one of $c$ classes. Suppose that we have a data set $\mathcal{D}$ = $\{d_{i},\mathbf{l_{i}}\}^{N}$, where $d_{i}$ denotes the text and the one-hot vector $\mathbf{l_{i}} \in \mathbb{R}^{c}$ represents the label for $d_{i}$, our goal is to learn a neural network $\mathcal{N}$ to classify the text.
\item Multi-Label Classification. 
In this task, each text instance belongs to a set of $c$ target labels. Formally, suppose that we have a dataset $\mathcal{D}$ = $\{d_{i},\mathbf{l_{i}}\}_{i=1}^{N}$, where $d_{i}$ denotes the text and the multi-hot vector $\mathbf{l_{i}}$ represents the label for the text $d_{i}$. Our goal is to learn a neural network $\mathcal{N}$ to classify the text. 
\end{itemize}

\begin{figure}[tbp] \centering  
	\includegraphics[width=0.95\columnwidth]{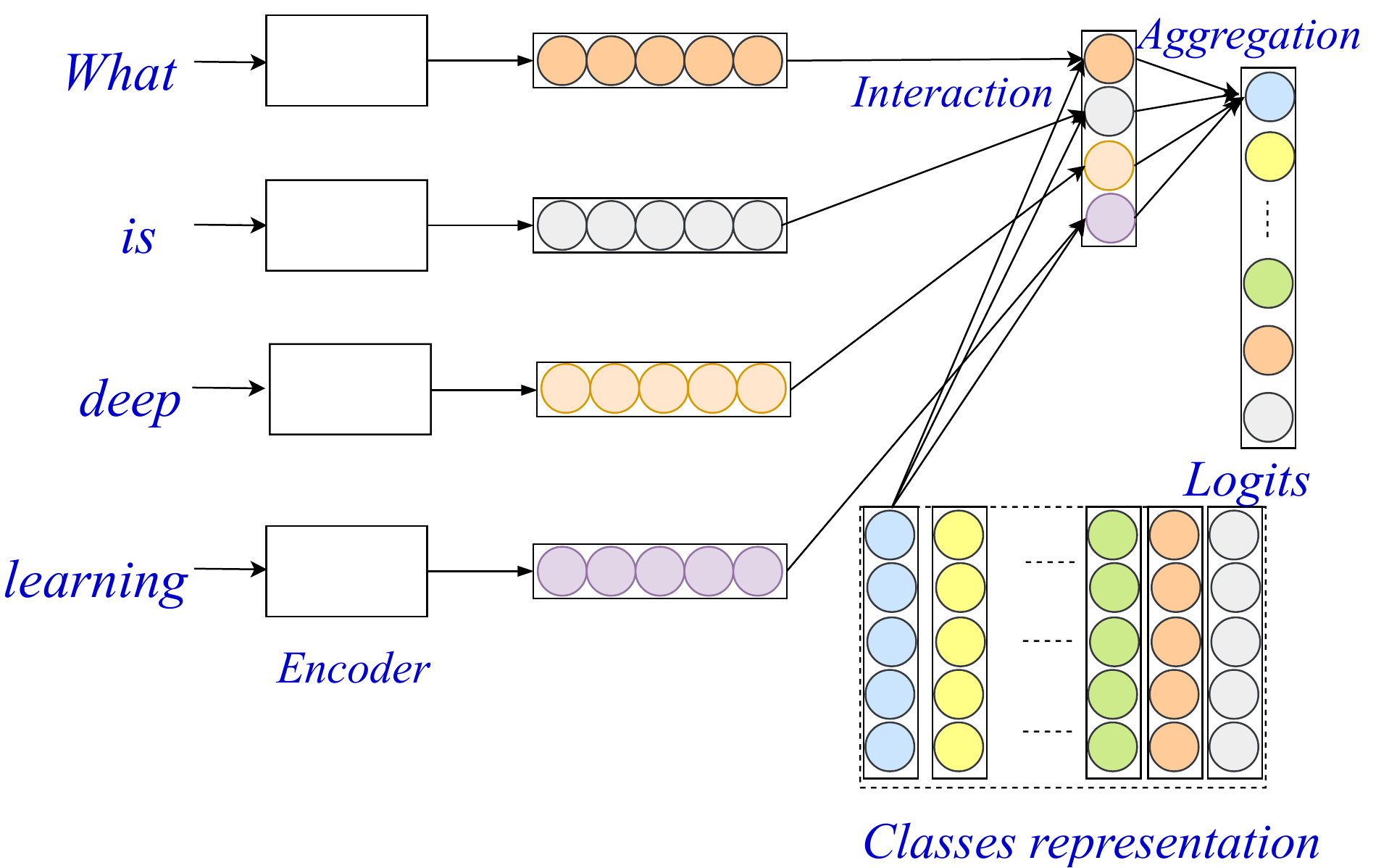}   
	\caption{Illustration of proposed EXAM method with word-level matching.} 
    	\label{exama}
\end{figure}
 
\subsection{Model Overview}
Motivated by the limitation of encoding-based models for text classification, which is lacking the fine-grained classification clue, we propose a novel framework, named \textit{EXplicit interAction Model} (EXAM), leveraging the interaction mechanism to incorporate word-level matching signals. As can be seen from Figure \ref{exama}, EXAM mainly contains three components:
\begin{itemize}
\item A \textit{word-level encoder} to project the input text $d_i$ into a word-level representation $\mathbf{H}$.
\item An \textit{interaction layer} to compute the interaction signals between the words and classes.
\item An \textit{aggregation layer} to aggregate the interaction signals for each class and make the final predictions.
\end{itemize}

Considering that word-level encoders are well investigated in previous studies (as mentioned in the Section 2), and the target of this work is to learn the fine-grained classification signals, we only elaborate the interaction layer and aggregation layer in the following subsections.

\begin{table*}
  \caption{Statistics of Datasets.}
  \label{multi-class}
  \centering
  \begin{tabular}{lcccccc}
    \toprule
    Dataset  &Classes&Average Lengths&Train Samples&Test Samples&Tasks\\
    \midrule
    Amazon Review Polarity	&2&91 	&3,600,000		&400,000& Sentiment\\
	Amazon Review Full		&5&93 		&3,000,000		&650,000&	 Analysis\\
    \midrule
	AG's News				&4&44 	&120,000		&7,600		&News Classification \\
    \midrule
	Yahoo! Answers			&10&112 &1,400,000		&60,000		&Question Answer \\
    \midrule
	DBPedia					&14&55 	&560,000		&70,000		&Ontology Extraction\\
   
    \bottomrule
  \end{tabular}
\end{table*}

\subsection{Interaction Layer}
\textit{Interaction mechanism} is widely used in tasks of matching source and target textual contents, such as natural language inference~\cite{llstm} and retrieve-based chatbot~\cite{smn}. The key idea of interaction mechanism is to use the interaction features between the small units (\eg words in the textual contents) to infer fine-grained clues whether two contents are matching. 
Inspired by the success of methods equipped with interaction mechanism over encode-based methods in matching the textual contents, we introduce the interaction mechanism into the task of matching textual contents with their classes (\ie text classification). 

Specifically, we devise an interaction layer which aims to compute the matching score between the word and class. Different from conventional interaction layer, where the word-level representations of both source and target are extracted with encoders like GRU, here we first project classes into real-valued latent representations. In other words, we employ a trainable representation matrix $\mathbf{T} \in \mathbb{R}^{c\times k}$ to encode classes (each row represents a class), where $c$ denotes the amount of classes and $k$ is the embedding size equals to that of words. 
We then adopt \textit{dot product} as the \textit{interaction function} to estimate the matching score between the target word $t$ and class $s$, of which the formulation is,
\begin{equation}
\mathbf{I_{st}} =\mathbf{T_{s,:}} \mathbf{H_{t,:}^{T}},
\end{equation}
where $\mathbf{H} \in \mathbb{R}^{n \times k}$ denotes word-level representation of the text, extracted by the encoder with $n$ denoting the length of the text. 
In this way, we can compute the interaction matrix $\mathbf{I} \in \mathbb{R}^{c\times n}$ by following:
\begin{equation}
\mathbf{I} =\mathbf{T} \mathbf{H^{T}}.
\end{equation}
Note that we reject more complex interaction functions like \textit{element-wise multiply} \cite{diin} and \textit{cosine similarity} \cite{BIMPM} for the consideration of efficiency.

\subsection{Aggregation Layer}
This layer is devised to aggregate the interaction features for each class $s$ into a logits $o_i^s$, which denotes the matching score between class $s$ and the input text $d_i$. The aggregation layer can be implemented in different ways such as CNN \cite{diin} and LSTM \cite{BIMPM}. However, to keep the simplicity and efficiency of EXAM, here we only use a MLP with two FC layers, where ReLU is employed as the activation function of the first layer. Formally, the MLP aggregates the interaction features $\mathbf{I_{s,:}}$ for class $s$, and compute its associated logits as following:
\begin{equation}
\left\{
\begin{array}{lr}
\mathbf{A_{s,:}} = ReLU(\mathbf{I_{s,:}} \mathbf{W_{1}} + \mathbf{b}),\\
o_i^{s} = \mathbf{A_{s,:}} \mathbf{W_{2}},\\
\end{array}
\right.
\end{equation}
where $\mathbf{W_{1}}$ and $\mathbf{W_{2}}$ are trainable parameters and $\mathbf{b}$ is the bias in the first layer.

We then normalize the logits $\mathbf{o_i} = [o_i^1, \cdots, o_i^c]$ into probabilities $\mathbf{p_i}$. Note that we follow previous work \cite{fasttext} and employ \textit{softmax} and \textit{sigmoid} for multi-class and multi-label classifications, respectively.

\subsection{Loss Function}
Similar to previous studies \cite{vdcnn}, in the multi-class text classification, we use cross entorpy loss as our loss function:
\begin{equation}
\mathcal{L}_{loss} =  -\sum_{i=1}^{N}\sum_{j=1}^{c}(l_{i}^{j}log(p_{i}^{j})). 
\end{equation}
Following previous researchers \cite{fasttext}, we choose binary classification loss as our loss function for the multi-label one:
\begin{equation}
\mathcal{L}_{loss} =  -\sum_{i=1}^{N}\sum_{j=1}^{c}(l_{i}^{j}log(p_{i}^{j})+(1-l_{i}^{j})log(1-p_{i}^{j})). 
\end{equation}

\section{Generalized Encoding-Based Model}
In this section, we elaborate how the \textit{encoding-based model} can be interpreted as a special case of our EXAM framework. As FastText \cite{fasttext} is the most popular model for text classification and has been investigated extensively in the literature, being able to recover it allows EXAM to mimic a large family of text classification models.

FastText contains three layers: 1) an embedding layer to get the word-level representation $\mathbf{H_{t,:}}$ for the word $t$, 2) an average pooling layer to get the text-level representation $\mathbf{f} \in \mathbb{R}^{1 \times k}$, and 3) a FC layer to get the final logits $\mathbf{p} \in \mathbb{R}^{1 \times c}$, where $k$ denotes the embedding size and $c$ means the number of classes. Note that we omit the subscript of the document ID for conciseness. Formally, it computes the logits $p^{s}$ of $s$-th class as follows:
\begin{equation}
\left\{
\begin{array}{lr}
\mathbf{f} = \frac{1}{n}\sum_{t=1}^{n}\mathbf{H_{t,:}},\\
p^{s} = \mathbf{fW_{:,s} + b_{s}},\\
\end{array}
\right.
\end{equation}
where $\mathbf{W} \in \mathbb{R}^{k \times c}$ and $\mathbf{b} \in \mathbb{R}^{1 \times c}$ are the trainable parameters in the last FC layer, and $n$ denotes the length of the text. The Eqn.(9) has an equivalent form as following:
\begin{equation}
p^{s} = \frac{1}{n}\sum_{t=1}^{n}\mathbf{(H_{t,:}W_{:,s}) + b_{s}}.\\
\end{equation}
It is worth noting that $\mathbf{H_{t,:}W_{:,s}}$ is exactly the interaction feature between word $t$ and class $s$. Therefore, the FastText is a special case of EXAM with an \textit{average pooling} as the aggregation layer. In EXAM, we use a non-linear MLP to be the aggregation layer, and it will generalize FastText to a non-linear setting which might be more expressive than the original one.
 \begin{table*}
  \caption{Test Set Accuracy [\%] on multi-class document classification tasks, and all the results of baselines are directly cited from the respective papers. The three different models are separated by lines. The best scores for the category are marked with underline and the overall best scores are highlight with bold font.}
  \label{multi-class}
  \centering
  \begin{tabular}{lcccccc}
    \toprule
    Model   &Amz. P. & Amz. F. &AG    &Yah. A. & DBP \\
    \midrule
    BoW \cite{charcnn}  & 90.4&\underline{54.6}  &88.8  &\underline{68.9} &96.6    \\
    N-grams \cite{charcnn}  &\underline{92.0} &54.3 &92.0  &68.5 &98.6    \\
    N-grams TFIDF \cite{charcnn} &91.5 & 52.4 &\underline{92.4}  &68.5 &\underline{98.7}  \\
    \midrule
    Char-CNN \cite{charcnn}  &94.5 & 59.6 &87.2  &71.2 &98.3    \\
    Char-CRNN \cite{charcnn}  &94.1 & 59.2 &\underline{91.4}  &71.7 &98.6      \\    
    VDCNN \cite{vdcnn} &\underline{\textbf{95.7}} & \underline{\textbf{63.0}} &91.3  &\underline{73.4} &\underline{98.7}    \\
    \midrule
    Small word CNN \cite{charcnn}  &94.2 & 56.3 &89.1 &70.0 &98.2      \\
    Large word CNN \cite{charcnn}  &94.2 & 54.1 &91.5  &71.0 &98.3      \\
    LSTM \cite{charcnn} &93.9 &59.4 &86.1 &70.8 &98.6\\
    Bigram-FastText \cite{fasttext} &94.6 &60.2 &92.5 &72.3 &98.6\\
    W.C RegionEmb \cite{regionemb} &95.1 &60.9 &92.8 &73.7 &98.9 \\
    EXAM (Ours)  &\underline{95.5} &\underline{61.9} & \underline{\textbf{93.0}}& \underline{\textbf{74.8}}& \underline{\textbf{99.0}}\\
    \bottomrule
  \end{tabular}
\end{table*}

\section{Experiments}
\subsection{Multi-Class Classification}
\subsubsection{Datasets}
We used publicly available benchmark datasets from \cite{charcnn} to evaluate EXAM. There are in total 6 text classification datasets, corresponding to sentiment analysis, news classification, question-answer and ontology extraction tasks, respectively. Table 1 shows the descriptive statistics of datasets used in our experiments. Stanford tokenizer is used to tokenize the text and all words are converted to lower case. We used padding to handle the various lengths of the text, and different maximum lengths are set for each dataset, respectively. If the length of the text is less than the corresponding predefined value, we padded it with zero; otherwise we truncated the original text. To guarantee a fair comparison, the same evaluation protocol of \cite{charcnn} is employed. We split 10\% samples from the training set as the validation set to perform early stop for our models.

\subsubsection{Hyperparameters}
For the multi-class task, we chose region embedding as the Encoder in EXAM. The region size is 7 and embedding size is 128. We used adam \cite{adam} as the optimizer with the initial learning rate 0.0001 and the batch size is set to 16. As for the aggregation MLP, we set the size of the hidden layer as 2 times interaction feature length. Our models are implemented and trained by MXNet \cite{MXNet} with a single NVIDIA TITAN Xp.
\subsubsection{Baselines}
To demonstrate the effectiveness of our proposed EXAM, we compared it with several state-of-the-art baselines. The baselines are mainly in three variants: 1) models based on feature engineering; 2) Char-based deep models, and 3) Word-based deep models. The first category uses the feature from the text to conduct the classification, and we reported the results from BoW \cite{charcnn}, N-grams \cite{charcnn} and N-grams TFIDF \cite{charcnn} as baselines. The second one means the input of the model is the character in the original text, and we chose the Char-CNN \cite{charcnn}, Char-CRNN \cite{charcnn} and VDCNN \cite{vdcnn} as baselines. As for the word-based deep models, the text is pre-segmented into words as the input, and we applied Small word CNN \cite{charcnn}, Large word CNN \cite{charcnn}, LSTM \cite{charcnn}, FastText \cite{fasttext} and W.C RegionEmb \cite{regionemb} as the baselines.  It is worth emphasizing that all the baselines and our EXAM do not use pre-trained word embedding over other corpus like glove.
\subsubsection{Overall Performance}
We compared our EXAM to several state-of-the-art baselines with respect to accuracy. All results are summarized in Table \ref{multi-class}. Four points are observed as following:
\begin{itemize}
\item Models based on feature engineering get the worst results on all the five datasets compared to the other methods. The main reason is that the feature engineering cannot take full advantage of the supervision from the training set and it also suffers from the data sparsity.
\item Char-based models get the highest overall scores on the two Amazon datasets. There are possibly two reasons, 1) compared to the word-based models, char-based models enrich the supervision from characters and the characters are combined to form N-grams, stems, words and phrase which are helpful in the sentimental classification. 2) The two Amazon datasets contain millions of training samples, perfectly fitting the deep residual architecture for the VDCNN. For the three char-based baselines, VDCNN gets the best performance on almost all the datasets because it has 29 convolutional layers allowing the model to learn more combinations of characters.
\item Word-based baselines exceed the other variants on three datasets and lose on the two Amazon datasets. The main reason is that the three tasks like news classification conduct categorization mainly via key words, and the word-based models are able to directly use the word embedding without combining the characters. For the five baselines, W.C RegionEmb performs the best, because it learns the region embedding to utilize the N-grams feature from the text.
\item It is clear to see that EXAM achieves the best performance over the three datasets: AG, Yah. A. and DBP. For the Yah.A., EXAM improves the best performance by 1.1\%. Additionally, as a word-based model, EXAM beats all the word-based baselines on the other two Amazon datasets with a performance gain of 1.0\% on the Amazon Full, because our EXAM considers more fine-grained interaction features between classes and words, which is quite helpful in this task.
\end{itemize}
\begin{table}
  \caption{Component-wise evaluation.}
  \label{componment}
  \centering
  \begin{tabular}{lcc}
    \toprule
     Dataset & EXAM & EXAM$_{Encoder}$\\
         \midrule
     Amz. P. & \textbf{95.5}  &95.1 \\
     Amz. F. & \textbf{61.9} &60.9 \\
     AG      &\textbf{93.0}  & 92.8\\
     Yah. A. &\textbf{74.8}  &73.1\\
     DBP     &\textbf{99.0} &98.9 \\
    \bottomrule
  \end{tabular}
\end{table}
\subsubsection{Component-wise Evaluation}
\begin{figure*}[!htb]

\centering
\subfigure{\includegraphics[scale=0.2]{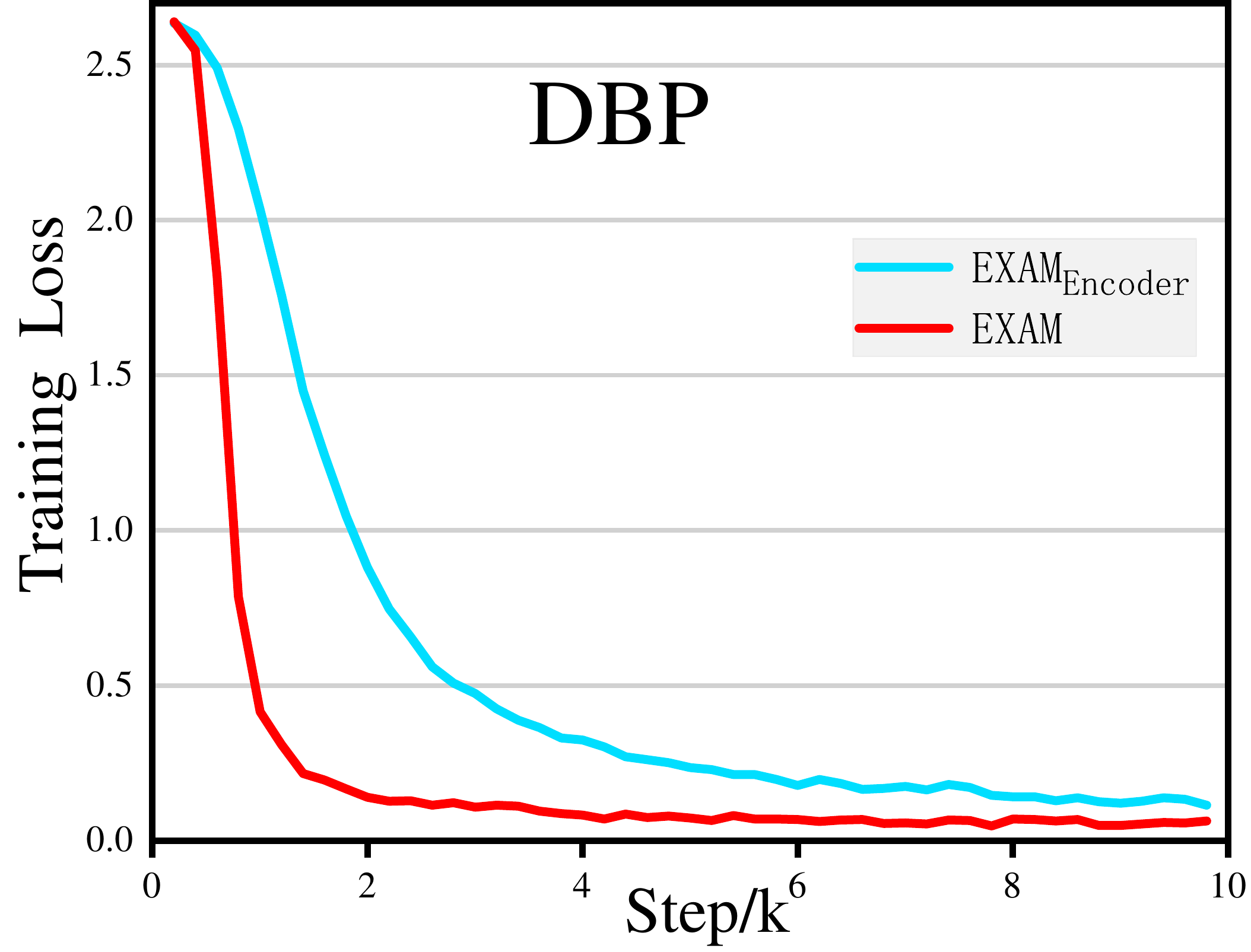}}
\subfigure{\includegraphics[scale=0.2]{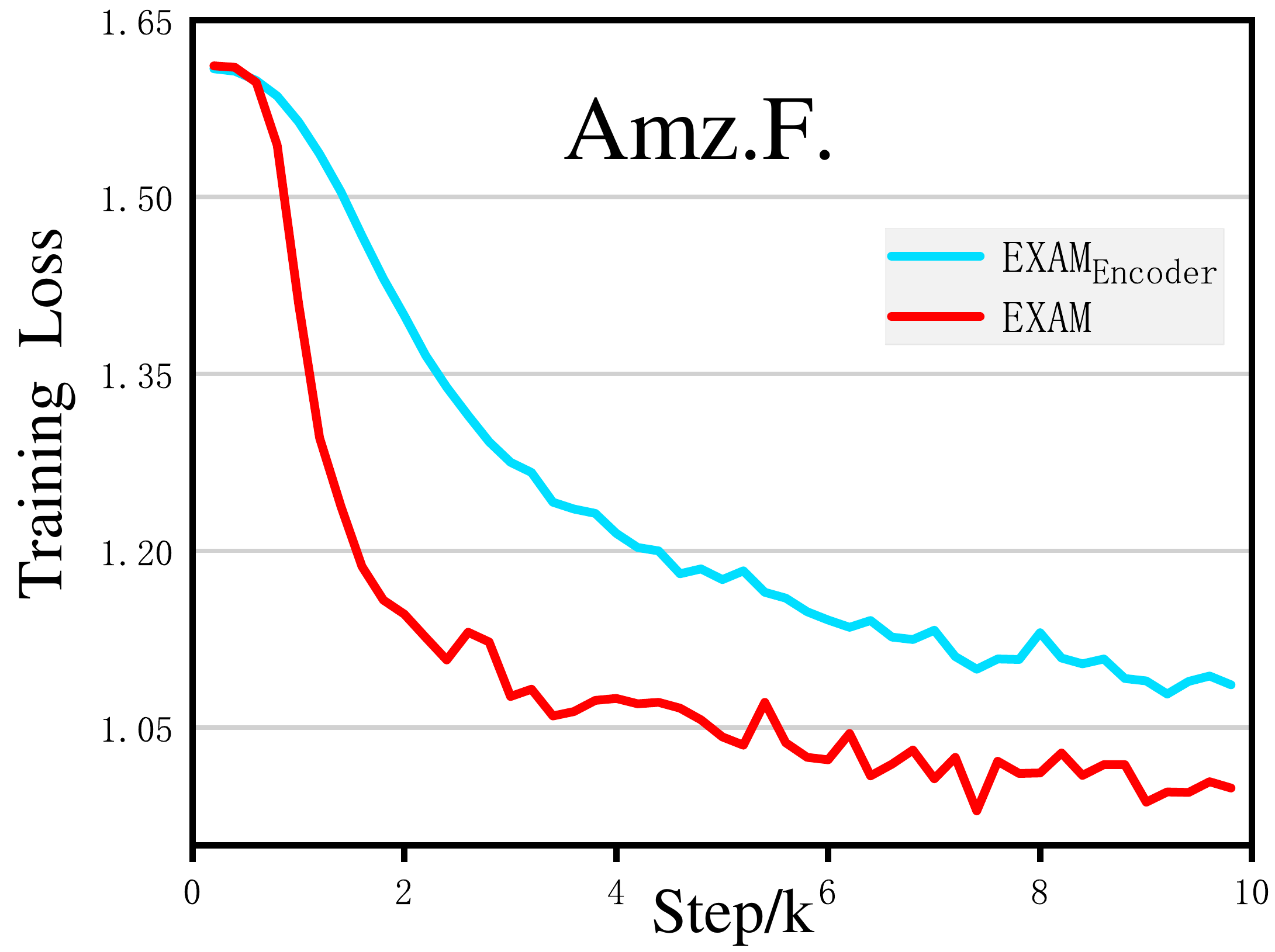}}
\subfigure{\includegraphics[scale=0.2]{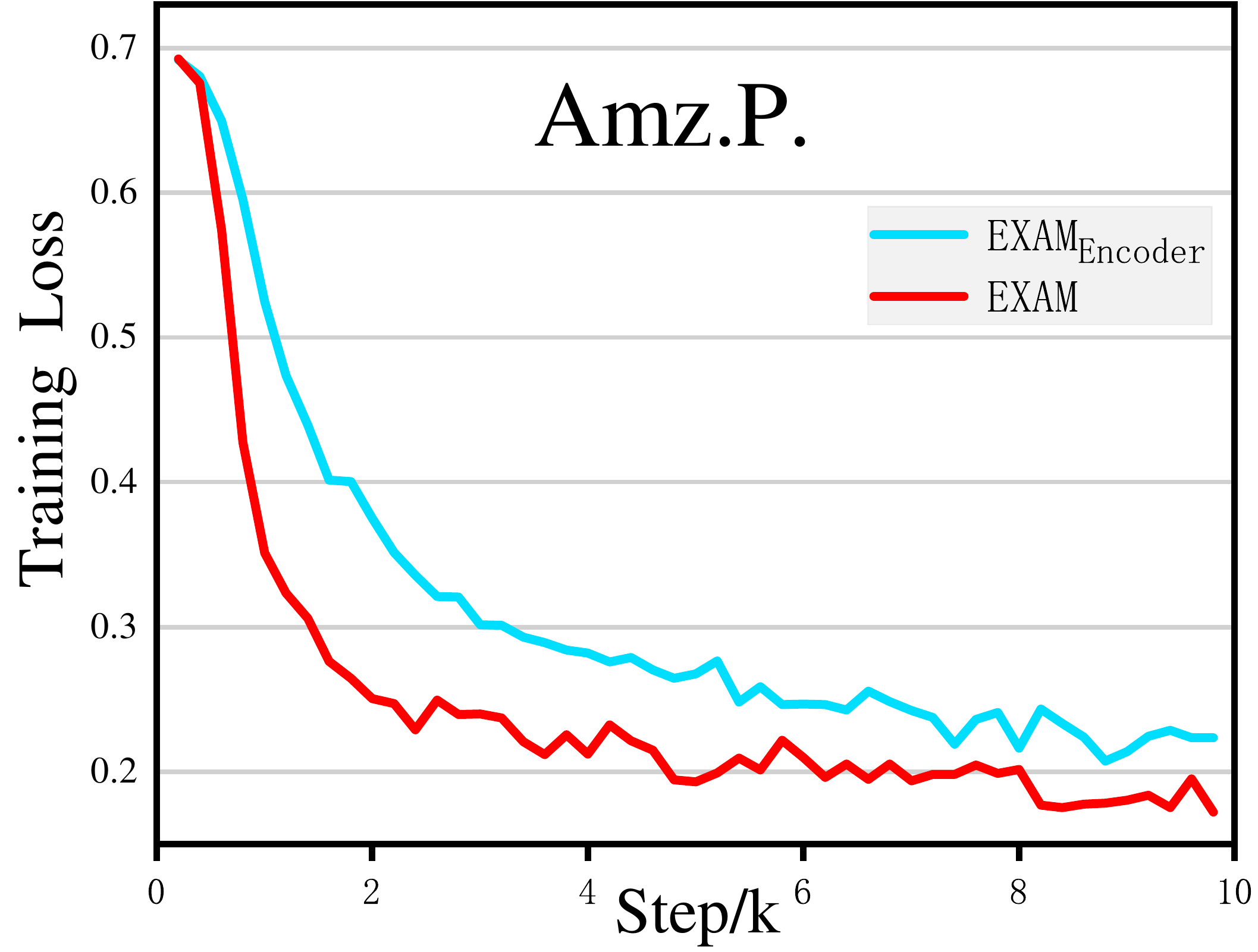}}
\subfigure{\includegraphics[scale=0.2]{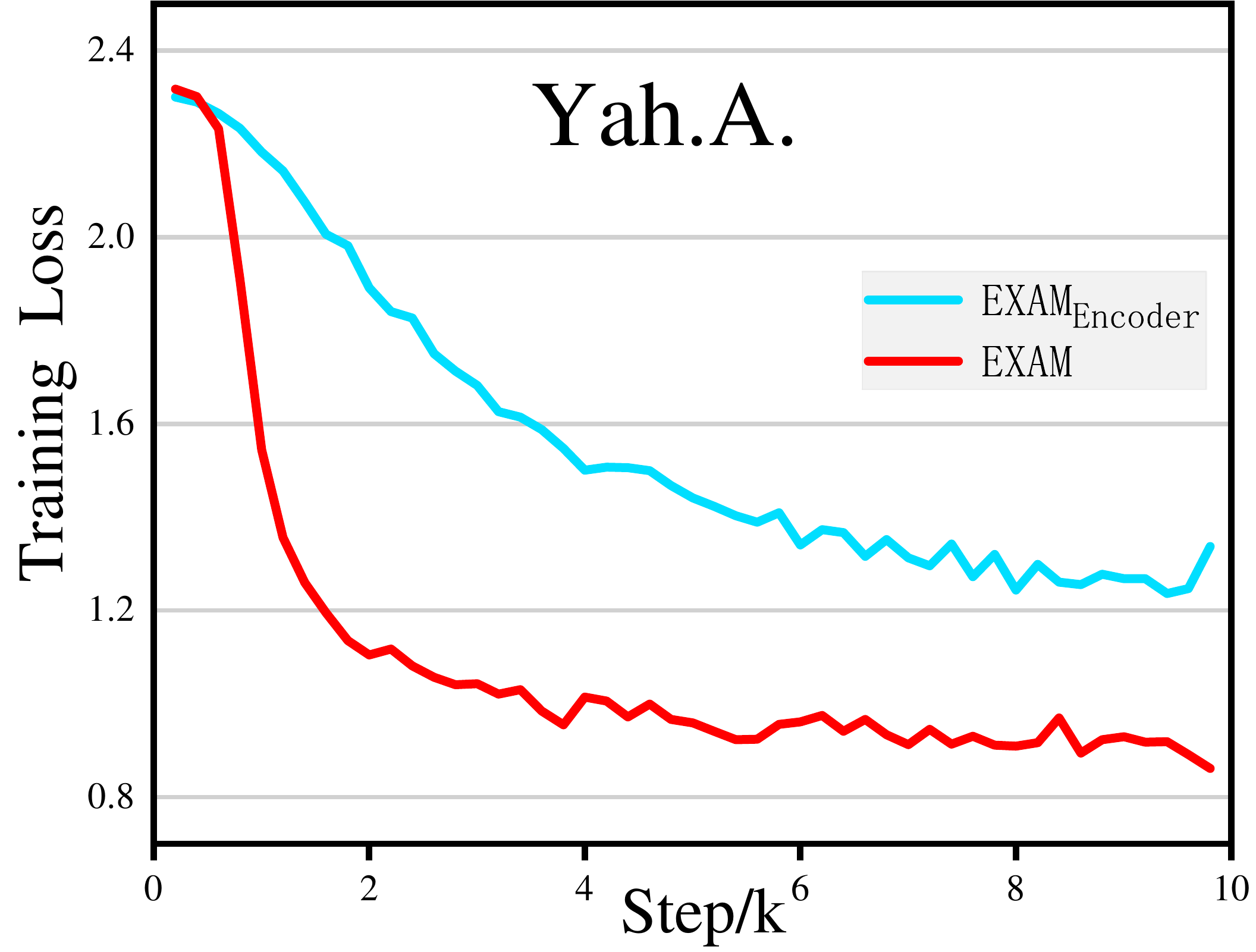}}
\caption{Convergence lines on the four dataset DBP, Amz. F., Amz. P. and Yah. A., respectively.}
\label{loss}
\end{figure*}

We studied the variant of our model to further investigate the effectiveness of the interaction layer and aggregation layer. We built a model called EXAM$_{Encoder}$ to preserve only the Encoder component with a max pooling layer and FC layer to derive the final probabilities. EXAM$_{Encoder}$ does not consider the interaction features between the classes and words, so it will automatically be degenerated into the Encoding-Based model. We reported the results of the two models on all the datasets at Table \ref{componment}, and it is clear to see that EXAM$_{Encoder}$ is not a patch on the original EXAM, verifying the effectiveness of interaction mechanism.
We also drew the convergence lines for EXAM and the EXAM$_{Encoder}$ for the datasets. From the Figure \ref{loss}, where the red lines represent EXAM and the blue is EXAM$_{Encoder}$, we observed that EXAM converges faster than EXAM$_{Encoder}$ with respect to all the datasets. Therefore, the interaction brings not only performance improvement but also faster convergence. The possible reason is that a non-linear aggregation layer introduces more parameters to fit the interaction features compared to the average pooling layer as mentioned in Section 4.

\begin{table*}
  \caption{Performance comparison between EXAM and baselines. The best scores are highlight in bold font.}
  \label{table}
  \centering
  \begin{tabular}{lcccccc}
    \toprule
     & \multicolumn{3}{c}{Kanshan-Cup Dataset} & \multicolumn{3}{c}{Zhihu Dataset} \\
    \cmidrule(r){2-4} \cmidrule(r){5-7}
    Model     &Precision & Recall@5 &$F_{1}$    &Precision & Recall@5 &$F_{1}$ \\
    \midrule
    Char-CNN \cite{charcnn} &1.299 & 0.536 &0.379  &- & - &-    \\
    Char-TextRNN \cite{TextRNN} &1.304 &0.537 &0.380  &- & - &-    \\
     \midrule
    FastText \cite{fasttext} &1.325 & 0.546 &0.387  &1.235 &0.564 &0.387 \\
    TextCNN \cite{TextCNN} &1.331 & 0.550 &0.389  &1.241 &0.566 &0.389    \\
    Word-TextRNN \cite{TextRNN} &1.345 & 0.555 &0.393  &1.240 &0.566  &0.389    \\
    EXAM (Ours) &\textbf{1.360} &\textbf{0.561} &\textbf{0.397} &\textbf{1.267}  &\textbf{0.578}  &\textbf{0.397}    \\
    \bottomrule
  \end{tabular}
\end{table*}
\begin{figure*}
\centering
\includegraphics[scale=0.5]{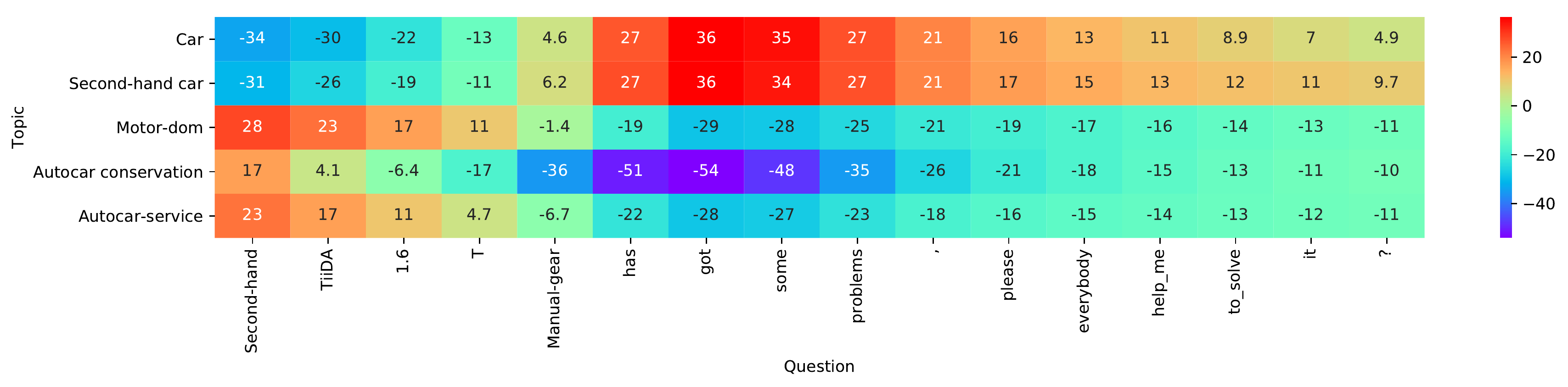}
\caption{The visualization of interaction features of EXAM.}
\label{vis4text}
\end{figure*}
\subsection{Multi-Label Classification}
\subsubsection{Datasets}
We conducted experiments on two different multi-label text classification datasets, named KanShan-Cup dataset\footnote{\url{https://biendata.com/competition/zhihu/.}} (a benchmark) and Zhihu dataset\footnote{\url{www.zhihu.com.}}, respectively. 
\begin{itemize}
\item \textbf{KanShan-Cup dataset}.
This dataset is released by a competition of tagging topics for questions (\textit{multi-label classification}) posted in the largest Chinese community question answering platform, Zhihu. The dataset contains 3,000,000 questions and 1,999 topics (classes), where one question may belong to one to five topics. For questions with more than 30 words, we kept the last 30 words, otherwise, we padded zeros. We separated the dataset into training, validation, and testing with 2,800,000, 20,000, and 180,000 questions, respectively.

\item \textbf{Zhihu dataset}.
Considering the user privacy and data security, KanShan-Cup does not provide the original texts of the questions and topics, but uses numbered codes and numbered segmented words to represent text messages. Therefore, it is inconvenient for researchers to perform analyses like visualization and case study. To solve this problem, we constructed a dataset named Zhihu dataset. We chose the top 1,999 frequent topics from Zhihu and crawled all the questions relevant to these topics. Finally, we acquired 3,300,000 questions, with less than 5 topics for each question. We adopted 3,000,000 samples as the training set, 30,000 samples as validation and 300,000 samples as testing.
\end{itemize}
\subsubsection{Baselines}
We applied the following models as baselines to evaluate the effectiveness of EXAM.
\begin{itemize}
\item \textbf{Char-based Model}. We chose Char-CNN \cite{charcnn} and Char-RNN \cite{TextRNN} as the baselines to represent this kind of methods. 
\item \textbf{Word-based Model}. For the word-based models, we reported the results from TextCNN \cite{TextCNN}, TextRNN \cite{TextRNN} and FastText \cite{fasttext}. The three models got the best performance in the KanShan-Cup competition, so we applied them as the word-based baselines.
\end{itemize}
\subsubsection{Hyperparameters}
We implemented the baseline models and EXAM by MXNet \cite{MXNet}. We used the matrix trained by word2vec \cite{Word2vec} to initialize the embedding layer, and the embedding size is 256. We adopted GRU as the Encoder, and each GRU Cell has 1,024 hidden states. The accumulated MLP has 60 hidden units. We applied Adam \cite{adam} to optimize models on one NVIDIA TITAN Xp with the batch size of 1000 and the initial learning rate is 0.001. The validation set is applied for early-stopping to avoid overfitting. All hyperparameters are chosen empirically.
\subsubsection{Metrics}
We used the following metrics to evaluate the performance of our model and baseline models.

\begin{itemize}
\item Precision: Different from the traditional precision metric (Precision@5) which is set as the fraction of the relevant topic tags among the five returned tags, we utilized weighted precision to encourage the relevant topic tags to be ranked higher in the returned list. Formally, the Precision is computed as following,
\begin{equation}
Precision = \sum\limits_{pos \in \{1,2,3,4,5\}} \frac{Precision @ pos}{log(pos + 1)}.
\end{equation}
\item Recall@5: Recall is the fraction of relevant topic tags that have been retrieved over the total amount of five relevant topic tags,  high recall means that the model returns most of the relevant topic tags. 
\item $F_{1}$: $F_{1}$ is the harmonic average of the precision and recall, we computed it as following,
\begin{equation}
F_{1} = \frac{Precision * Recall@5}{Precision + Recall@5}. 
\end{equation}
\end{itemize}

\subsubsection{Performance Comparison}

Table \ref{table} gives the performance of our model and baselines over two different datasets with respect to Precision, Recall@5 and $F_{1}$. We observed the following from the Table \ref{table}:
\begin{itemize}
\item Word-based models are better than char-based models in Kanshan-Cup dataset. That may be because in Chinese the words can offer more supervisions than characters and the question tagging task needs more word supervision.
\item  For word-based baseline models, all the baselines have similar performance which corroborates the conclusion in FastText \cite{fasttext} that simple network is on par with deep learning classifiers in text classification.
\item Our models achieve the state-of-the-art performance over two different datasets though we only slightly modified TextRNN to build EXAM. Different from the traditional models which encode the whole text into a vector, in EXAM, the representations of classes firstly interact with  words to get more fine-grained features as shown in Figure \ref{vis4text}. The results suggest that word-level interaction features are relatively more important than global text-level representations in this task.
\end{itemize}

\subsubsection{Interaction Visualization}
To illustrate the effectiveness of explicit interaction, we visualized an interaction feature \textbf{I} of the question ``Second-hand TIDDA 1.6 T Mannual gear has gotten some problems, please everybody help me to solve it ?''. This question has 5 topics: Car, Second-hand Car, Motor Dom, Autocar Conversation and Autocar Service. EXAM only misclassified the last topic. In  Figure \ref{vis4text}, we observed that  when classifying different topics, the interaction features are different. The topics ``Car'' and ``Second-hand Car'' pay much attention to the words like ``Second-hand TIIDA'' and the other topic like ``Autocar Conversation'' focuses more on  ``got some problems''. The results clearly signify that the interaction feature between the word and class is well-learned and highly meaningful.

\section{Related Work}

\subsubsection{Text Classification}

Existing researches on text classification can be categorized into two groups: feature-based and deep neural models. The former focuses on hand-craft features and uses machine learning algorithms as the classifier. Bag-of-words \cite{bow} is a very efficient way to conduct the feature engineering. SVM and Naive Bayes are constantly the classifier. The latter, deep neural models, taking advantage of neural networks to accomplish the model learning from data, have become the promising solution for the text classification. For instance, Iyyer et al. \shortcite{dan} proposed Deep Averaging Networks (DAN) and Grave et al. \shortcite{fasttext} proposed the FastText, and both are simple but efficient. To get the temporal features between the words in the text, some models like TextCNN \cite{TextCNN} and Char-CNN \cite{charcnn} exploit the convolutional neural network, and there are also some models based on Recurrent Neural Network (RNN). Recently, Johnson et al. \shortcite{vdcnn} investigated the residual architecture and built a model called VD-CNN and Qiao et al. \shortcite{regionemb} proposed a new method of region embedding for the text classification. However, as mentioned in the Introduction, all these methods are text-level models while EXAM conducts the matching at the word level.
\subsubsection{Interaction Mechanism}
Interaction Mechanism is widely used in Natural Language Sentence Matching (NLSM). The key idea of interaction mechanism is to use the interaction features between the small units (like words in sentence) to make the matching. Wang et al. \shortcite{llstm} proposed a ``matching-aggregation'' framework to perform the interaction in Natural Language Inference. Following this work, Parikh et al. \shortcite{DATTENTION} integrated the attention mechanism into this framework, called Decomposable Attention Model. Then Wang et al. \shortcite{Wang2016ACM} discussed different interaction functions in Text Matching. Yu et al. \shortcite{TREE} adopted tree-LSTM to get different level units to perform the interaction. 
Gong et al. \shortcite{diin} proposed a densely interactive inference network to use DenseNet to aggregate dense interaction features. Our work is different from them since they mainly apply this mechanism in text matching instead of the classification. 

\section{Conclusion}
In this work, we present a novel framework named EXAM which employs the interaction mechanism to explicitly compute the word-level interaction signals for the text classification. We apply the proposed EXAM on multi-class and multi-label text classifications. Experiments over several benchmark datasets verify the effectiveness of our proposed mechanism. In the future, we plan to investigate the effect of different interaction functions in the interaction mechanism. Besides, we are interested in extend EXAM by introducing more complex aggregation layers like ResNet or DenseNet.

\section{ Acknowledgments}
This work is supported by the National Basic Research Program of China (973 Program), No.: 2015CB352502; National Natural Science Foundation of China, No.: 61772310, No.:61702300, and No.:61702302; the Project of Thousand Youth Talents 2016; and the Tencent AI Lab Rhino-Bird Joint Research Program (No.JR201805); Fundamental Research Funds of Shandong University (No. 2017HW001).

\bibliographystyle{aaai}
\bibliography{aaai}

\end{document}